\documentclass[sigconf]{acmart}
\AtBeginDocument{%
  }

\copyrightyear{2025}
\acmYear{2025}
\setcopyright{cc}
\setcctype{by-nd}
\acmConference[MM '25]{Proceedings of the 33rd ACM International Conference on Multimedia}{October 27--31, 2025}{Dublin, Ireland}
\acmBooktitle{Proceedings of the 33rd ACM International Conference on Multimedia (MM '25), October 27--31, 2025, Dublin, Ireland}
\acmDOI{10.1145/3746027.3754577}
\acmISBN{979-8-4007-2035-2/2025/10}

\acmSubmissionID{mfp7205}

\usepackage{subfigure}
\usepackage{multirow}
\usepackage{amsmath}
\usepackage{pifont}
\newcommand{\cmark}{\ding{51}}  
\newcommand{\xmark}{\ding{55}}  

\begin{document}

\title{AlignDiT: Multimodal Aligned Diffusion Transformer\\for Synchronized Speech Generation}
\renewcommand{\shorttitle}{AlignDiT: Multimodal Aligned Diffusion Transformer for Synchronized Speech Generation}

\author{Jeongsoo Choi}
\orcid{0009-0005-6817-604X}
\affiliation{%
  \institution{Korea Advanced Institute of Science and Technology}
  \city{Daejeon}
  \country{Republic of Korea}
}
\email{jeongsoo.choi@kaist.ac.kr}

\author{Ji-Hoon Kim}
\orcid{0009-0001-3433-918X}
\affiliation{%
  \institution{Korea Advanced Institute of Science and Technology}
  \city{Daejeon}
  \country{Republic of Korea}
}
\email{jihoon@mm.kaist.ac.kr}

\author{Kim Sung-Bin}
\orcid{0000-0003-3542-9934}
\affiliation{%
  \institution{Pohang University of Science and Technology}
  \city{Pohang}
  \country{Republic of Korea}
}
\email{sungbin@postech.ac.kr}

\author{Tae-Hyun Oh}
\orcid{0000-0003-0468-1571}
\affiliation{%
  \institution{Korea Advanced Institute of Science and Technology}
  \city{Daejeon}
  \country{Republic of Korea}
}
\email{taehyun.oh@kaist.ac.kr}

\author{Joon Son Chung}
\orcid{0000-0001-7741-7275}
\affiliation{%
  \institution{Korea Advanced Institute of Science and Technology}
  \city{Daejeon}
  \country{Republic of Korea}
}
\email{joonson@kaist.ac.kr}


\begin{abstract}
In this paper, we address the task of multimodal-to-speech generation, which aims to synthesize high-quality speech from multiple input modalities: text, video, and reference audio. This task has gained increasing attention due to its wide range of applications, such as film production, dubbing, and virtual avatars. Despite recent progress, existing methods still suffer from limitations in speech intelligibility, audio-video synchronization, speech naturalness, and voice similarity to the reference speaker. To address these challenges, we propose AlignDiT, a multimodal Aligned Diffusion Transformer that generates accurate, synchronized, and natural-sounding speech from aligned multimodal inputs. Built upon the in-context learning capability of the DiT architecture, AlignDiT explores three effective strategies to align multimodal representations. Furthermore, we introduce a novel multimodal classifier-free guidance mechanism that allows the model to adaptively balance information from each modality during speech synthesis. Extensive experiments demonstrate that AlignDiT significantly outperforms existing methods across multiple benchmarks in terms of quality, synchronization, and speaker similarity. Moreover, AlignDiT exhibits strong generalization capability across various multimodal tasks, such as video-to-speech synthesis and visual forced alignment, consistently achieving state-of-the-art performance. The demo page is available at \href{https://mm.kaist.ac.kr/projects/AlignDiT}{https://mm.kaist.ac.kr/projects/AlignDiT}.
\end{abstract}


\begin{teaserfigure}
  \includegraphics[width=\textwidth]{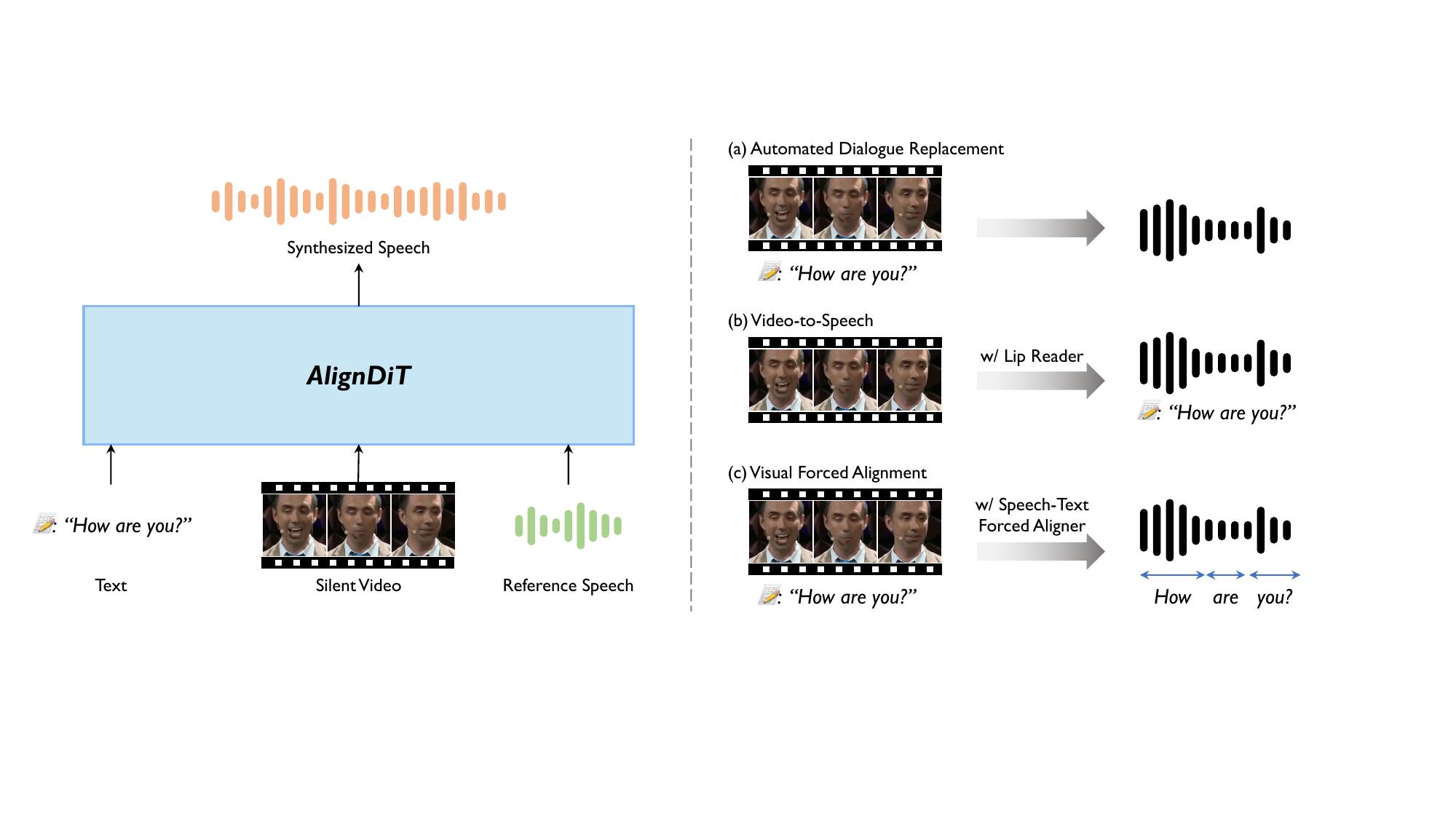}
  \vspace{-5mm}
  \caption{
  AlignDiT aims to generate natural speech that is (1) accurately aligned with the given text, (2) temporally synchronized with the input video, and (3) acoustically consistent with the reference speech.
  This versatile system has a broad range of applications, including (a) Automated dialogue replacement, (b) Video-to-Speech synthesis, and (c) Visual Forced Alignment.
  }
  \label{fig:teaser}
  \vspace{3mm}
\end{teaserfigure}

\maketitle

\section{Introduction}
Human communication involves the exchange of information through spoken language, which serves as a fundamental means of interaction between individuals~\cite{mcgurk1976hearing, holler2019multimodal, drijvers2023multimodal}. It naturally generates multimodal signals that convey both linguistic content and paralinguistic cues. Among them, audio, video, and text are three key modalities widely utilized in speech and language processing. The audio modality delivers the sound signal that carries phonetic content, prosodic features such as intonation and rhythm, as well as speaker-specific characteristics. The video modality captures lip movements, facial expressions, and non-verbal signals that are synchronized with the audio. Text, while not directly observable in the physical world, serves as a symbolic and discrete representation of the linguistic content, monotonically aligned with audio and video over time~\cite{petridis2018audio, ren2019fastspeech}. Together, these modalities complement one another, offering a comprehensive view of human verbal communication.

Cross-modal generation tasks, where one modality is synthesized or inferred from others, have been extensively studied to support accessibility and multimodal interaction. Typical examples include text-to-speech synthesis (TTS)~\cite{ren2021fastspeech, chen2024vall, chen2025f5, mehta2024matcha}, automatic speech recognition (ASR)~\cite{schneider2019wav2vec, baevski2020wav2vec, gulati2020conformer, radford2023robust}, lip reading~\cite{son2017lip, ma2021lip, ma2023auto, yeo2024akvsr}, video-to-speech synthesis~\cite{mira2022svts, yemini2024lipvoicer, kim2024let, choi2025v2sflow}, and talking face generation~\cite{prajwal2020lip, park2022synctalkface, xu2024vasa}. While many of these tasks focus on converting a single input modality into another, incorporating multiple input modalities can lead to more accurate and robust generation, as different modalities provide complementary information. For example, audio-visual speech recognition (AVSR)~\cite{afouras2018deep, hong2023watch, choi2024av2av} improves transcription accuracy by leveraging both audio and video inputs, especially under noisy conditions.

This motivates generating speech audio conditioned on both video and text inputs, where visual cues from lip movements and explicit linguistic content complement each other. A representative application of this task is Automated Dialogue Replacement (ADR)~\cite{cong2023learning, cong2024styledubber}, widely used in film and television post-production. During this process, actors' lines are re-recorded to improve audio quality while ensuring synchronization with the video, especially when original recordings are affected by background noise or challenging recording conditions. Building on recent advances in deep learning based speech generative models, prior studies have explored generating speech from silent video, transcriptions, and short reference audio clips. Despite their potential, existing multimodal-to-speech generation approaches face three key challenges. First, they often struggle to generate natural and intelligible speech due to limited modeling capacity and reliance on limited datasets. Second, they are not robust when one or more input modalities are missing or corrupted, as they lack mechanisms to adjust the importance of each modality. Third, many methods depend on external forced aligners or duration predictors for synchronization, increasing supervision costs and risks propagating alignment errors.

To address these challenges, we propose AlignDiT, a multimodal aligned diffusion transformer architecture designed for natural and synchronized speech generation. AlignDiT jointly models video, text, and reference audio within a unified framework, implicitly learning cross-modal alignments without relying on explicit duration predictors or external forced aligners. By framing speech synthesis as a conditional generative diffusion process, AlignDiT naturally aligns the generated speech with visual lip movements, linguistic content, and speaker-specific voice characteristics. We conduct extensive experiments using both subjective and objective evaluation metrics. The results clearly show that AlignDiT significantly outperforms existing ADR methods across all criteria, including speech intelligibility, synchronization accuracy, and speaker similarity. Furthermore, AlignDiT effectively generalizes to related multimodal tasks, such as video-to-speech synthesis and visual forced alignment, highlighting its robustness and flexibility across diverse multimodal scenarios, as demonstrated in Fig.~\ref{fig:teaser}.

Our major contributions can be summarized as follows:
\begin{itemize}
    \item We propose AlignDiT, a model that jointly leverages video, text, and reference audio to synthesize accurate, high-quality, and synchronized speech for the ADR task.
    \item We conduct extensive analyses and experiments to explore various settings and identify the most effective approach for multimodal alignment.
    \item We demonstrate the versatility of AlignDiT by successfully adapting it to related multimodal tasks, such as video-to-speech and visual forced alignment.
\end{itemize}

\begin{figure*}
    \centering
    \subfigure[Early Fusion \& Self-attention]{
        \includegraphics[height=5.7cm]{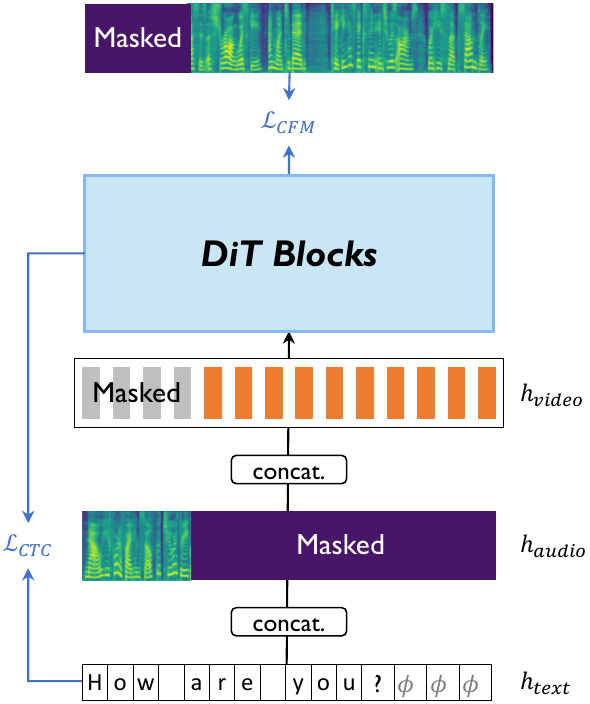}
        \label{fig:main1}}
    \subfigure[Prefix \& Self-attention]{
        \includegraphics[height=5.7cm]{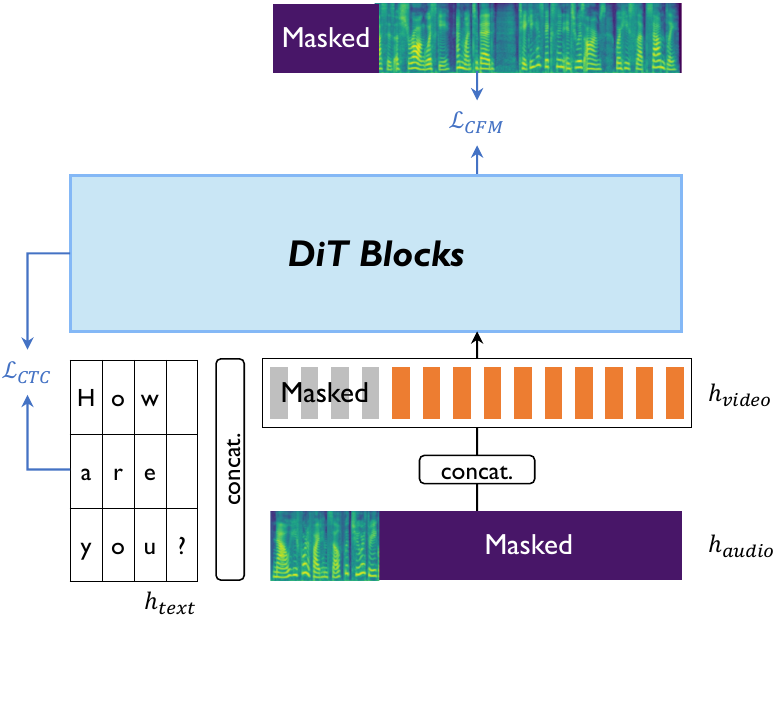}
        \label{fig:main2}}
    \subfigure[Multimodal Cross-attention]{
        \includegraphics[height=5.7cm]{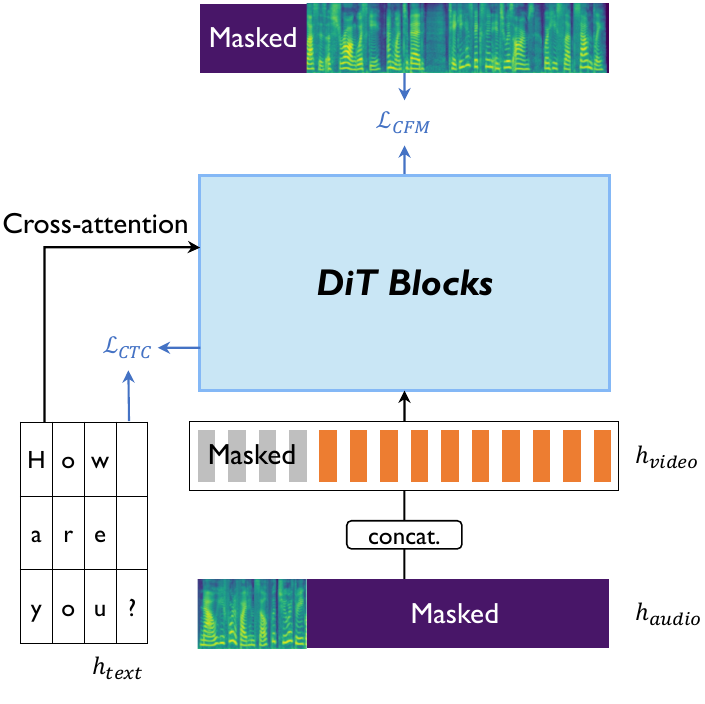}
        \label{fig:main3}}
    \vspace{-0.3cm}
    \caption{
    Various methods for conditioning multimodal inputs to DiT blocks: (a) channel-wise concatenation of text, reference speech, and visual features; (b) providing the text as a prefix; and (c) conditioning text inputs through cross-attention.
    In (a), \text{$\phi$} represents filler tokens.
    }
    \label{fig:main}
    \vspace{-0.3cm}
\end{figure*}

\section{Related Works}
\subsection{Multimodal Speech Tasks}
\noindent\textbf{Automated dialogue replacement (ADR).} Early efforts in ADR, also known as automated video dubbing, framed it as a multimodal text-to-speech (TTS) problem. Neural Dubber~\cite{hu2021neural} pioneered this direction by generating speech from text conditioned on video lip movements. Subsequent work, VisualTTS~\cite{lu2022visualtts}, integrated visual information more explicitly into the TTS pipeline by introducing a textual-visual attention mechanism to learn alignments between phonemes and lip frames, as well as visual feature fusion during acoustic decoding. VDTTS~\cite{hassid2022more} extended this idea to unconstrained, multi-speaker settings, predicting finer prosodic elements for a more natural dubbing. More recently, HPMDubbing~\cite{cong2023learning} presented a unified architecture employing hierarchical prosody modeling. It extracts visual features at the lip, face, and scene levels to control different aspects of speech (timing, energy/pitch, and global emotion), synthesizing natural and emotional speech and setting a strong state-of-the-art benchmark for ADR. StyleDubber~\cite{cong2024styledubber} further focused on speaker style and pronunciation habits by introducing two-level style learning (phoneme-level and utterance-level) using reference audio. Both HPMDubbing and StyleDubber require explicit alignment through a forced aligner during training to ensure accurate synchronization between video and speech.

\noindent\textbf{Video-to-speech.} Distinct from ADR, video-to-speech generates speech directly from silent videos without transcripts. Initial methods approached this task using CNN-based~\cite{ephrat2017vid2speech, kumar2019lipper} or sequence-to-sequence models~\cite{prajwal2020learning} to capture lip movements and generate speech. Recent approaches have further improved speech generation quality by incorporating advanced techniques, such as GANs~\cite{kim2021lip, mira2022end}, normalizing flows~\cite{he2022flow, kim2024let}, and diffusion models~\cite{choi2023diffv2s, yemini2024lipvoicer}. Additionally, to capture speaker characteristics, most works~\cite{choi2023intelligible, mira2022svts} utilize speaker embeddings derived from reference audio. As obtaining reference audio during inference is not always feasible, several studies~\cite{choi2023diffv2s, yemini2024lipvoicer, kim2024let, kim2025faces} extract speaker information directly from the given video to facilitate speaker-aware speech synthesis.

\noindent\textbf{Visual forced alignment (VFA).}
VFA is a task that identifies the timeline—specifically, the start and end times—for each word or phoneme in silent videos based on corresponding textual content. VFA requires accurately aligning lip movements with text. One approach employs visual keyword spotting~\cite{prajwal2021visual, momeni2020seeing}, a technique identifying durations of individual words. Repeating this across an entire video achieves visual forced alignment, but involves substantial computational demands and potential inaccuracies due to overlaps between adjacent words. Another common approach~\cite{kurzinger2020ctc} utilizes Connectionist Temporal Classification (CTC)~\cite{graves2006connectionist}, frequently employed in visual speech recognition (VSR) for video-text alignment. DVFA~\cite{kim2023deep} represents the first method explicitly designed for VFA, leveraging a multimodal attention mechanism to effectively align textual transcriptions with corresponding lip movements.

Our approach builds on insights from these three different lines of work, aiming to achieve the best of each by learning cross-modal alignment across speech, video, and text. Crucially, we avoid dedicated speech-text aligners, such as those required by HPMDubbing and StyleDubber, by learning implicit multimodal alignment. This results in improved content accuracy and speaker similarity. Furthermore, the flexibility of our multimodal alignment approach allows it to generalize beyond ADR, effectively tackling video-to-speech and VFA, thus demonstrating significant versatility and broad applicability.

\subsection{Text-to-speech (TTS) Synthesis}
Recent research in Text-to-Speech (TTS) synthesis has achieved significant advancements. One notable approach, autoregressive (AR)-based TTS models~\cite{wang2023neural, chen2024vall, kharitonov2023speak}, combines powerful speech tokenizers~\cite{defossez2023high,zeghidour2021soundstream} with next-token prediction language modeling and has demonstrated promising results. Despite their high quality, non-autoregressive (NAR) models benefit from fast inference due to parallel processing, effectively balancing quality and latency. Specifically, diffusion models~\cite{ho2020denoising,song2020score} have significantly contributed to the success of current NAR approaches. For example, Matcha-TTS~\cite{mehta2024matcha} adopts conditional flow matching with optimal transport paths (OT-CFM)~\cite{lipman2023flow} for training and relies on a phoneme-level duration model for speech synthesis. DiTTo-TTS~\cite{lee2025ditto} improves alignment by utilizing a Diffusion Transformer (DiT)~\cite{peebles2023scalable} with cross-attention conditioned on encoded text from a pretrained language model. E2 TTS~\cite{eskimez2024e2} removes phoneme and duration predictors, directly using characters padded with filler tokens to match the length of mel spectrograms. Additionally, F5-TTS~\cite{chen2025f5} enhances text-speech alignment by integrating ConvNeXt V2~\cite{woo2023convnext} into a diffusion transformer framework. In this work, we aim to extend NAR TTS models, specifically DiT, to multimodal TTS synthesis. By leveraging classifier-free guidance and carefully designed conditioning, our approach flexibly handles varying inputs—including text-only, video-only, and multimodal scenarios—for high-quality speech synthesis.

\section{AlignDiT}
Our goal is to generate a speech waveform that matches the lip movements in a video, conveys the content of a given text script, and resembles the voice characteristics of a target speaker indicated by reference speech. We employ an in-context learning-based speech synthesis approach to handle multimodal inputs. For text-to-speech synthesis task, various methods~\cite{eskimez2024e2, chen2025f5, yang2024simplespeech, lee2025ditto} have explored diverse strategies for conditioning text inputs. However, to the best of our knowledge, no existing method simultaneously handles video input alongside text and reference speech to achieve multimodal alignment for speech synthesis within in-context learning-based generative models. We aim to explore various settings and identify the most suitable method for multimodal alignment, capable of flexibly handling both unimodal and multimodal inputs to synthesize high-quality and accurate speech.

\subsection{Model Architecture}
We utilize the Diffusion Transformer (DiT)~\cite{peebles2023scalable} for multimodal-to-speech generation, as it has been shown to be effective for speech generation task. Following prior works about text-to-speech~\cite{le2023voicebox, chen2025f5}, we train the model using a flow matching objective and generate mel-spectrogram through an iterative inference process starting from random noise. The model consists of multiple blocks, each following the standard Transformer~\cite{vaswani2017attention} architecture with self-attention, along with additional parameters for conditioning diffusion timestep information. For multimodal-to-speech generation, we guide the generative process using a fused multimodal representation as a condition, allowing the model to progressively refine the noisy mel-spectrogram into a representation that aligns with the given multimodal inputs.

\subsection{Multimodal Conditioning}
\noindent\textbf{Audio-Video Fusion.} Audio and video are naturally synchronized modalities. When a person speaks, their lip movements correspond closely to the acoustic features at that same moment, making frame-by-frame fusion straightforward. However, since the frame rates of the two modalities typically differ, we extract and encode each modality to have a common temporal resolution. Specifically, audio is converted into mel‐spectrogram sequence $h_{audio}$ at 100~fps. For video, we first extract a sequence of video features $h_{video}$ using a pretrained video encoder specialized in lip motion. To match the frame rate, we upsample the 25~fps video features to 100~fps via lightweight transposed convolutional layers. We then apply Conformer~\cite{gulati2020conformer} encoder for better capturing contextual information. Once aligned temporally, the audio and video features are concatenated channel‐wise to form a unified multimodal representation, which effectively encodes when and how to speak.

To enable our model to generate speech that follows the voice characteristics of a reference speech, we apply a binary temporal mask $M$ to the mel-spectrogram and train the model to inpaint the masked regions, using $(1-M) \odot h_{audio}$ as input. The masked spans are randomly selected to enhance the in-context learning ability of the model. Since requiring paired audio-video data of reference utterance during inference is critical, we train the model to operate with reference speech alone by applying complementary masking to the video features. Specifically, the input becomes $M \odot h_{video}$, allowing the model to rely on audio while ignoring the masked video during training. This audio-video fusion can be formulated as follows:
\begin{equation}
h_{av}=[(1-M) \odot h_{audio}; M \odot h_{video}] \in \mathbb{R}^{T \times 2D},
\end{equation}
where $T$ and $D$ denote the temporal length and the hidden dimension, respectively.

\noindent\textbf{Audio-Video-Text Fusion.}
While the audio‐video streams are inherently time-aligned, text is only monotonically aligned with them and lacks strict frame-level synchronization. Unlike previous methods that rely on text token-level duration information from external forced aligners~\cite{mcauliffe2017montreal} and duration predictors~\cite{ren2021fastspeech}, we aim to train the generative model fuse the modalities naturally, without explicit duration constraints. This simplifies the data preprocessing pipeline, making it easier to scale the dataset, and avoids potential biases introduced by forced alignments, thereby facilitating a more natural audio-video-text alignment. To investigate feasible modality fusion methods in the DiT blocks, we explore three conditioning strategies as illustrated in Fig.~\ref{fig:main}. In all cases, for text encoding, the character sequence is first embedded through a lookup table and then refined by a convolutional encoder to be $h_{text}$ which has the length of $L$.

(a) Early Fusion \& Self‐Attention: 
A naive approach for fusion is concatenating all conditioning modalities along the channel axis. To match the total length, we add filler tokens at the end of the text sequence inspired by E2 TTS~\cite{eskimez2024e2}. After this early fusion, the self-attention layers in DiT blocks learn alignment naturally to predict the masked part of mel-spectrogram. This can be expressed as follows:
\begin{equation}
h'=[h_{av}; h_{text}] \in \mathbb{R}^{T \times 3D},
\end{equation}
\begin{equation}
h=h'W_a+b_a\in \mathbb{R}^{T \times D},
\end{equation}
where $W_a$ and $b_a$ are the parameters of a fully connected layer to transform the channel dimension to $D$.

(b) Prefix \& Self‐Attention:
Another approach for fusion is frame-wise concatenation between text feature and audio-video feature, treating the text as a prefix. This leverages in-context conditioning~\cite{wang2023neural, yang2024simplespeech}, allowing the text prefix to guide the model within the self-attention layers of DiT blocks. After the model outputs the full sequence, the prefix part corresponding to the text length is discarded, yielding the predicted mel-spectrogram. This can be formulated as follows:
\begin{equation}
h'=\text{Concat}(h_{text}, h_{av}) \in \mathbb{R}^{(L + T) \times 2D},
\end{equation}
\begin{equation}
h=h'W_b+b_b\in \mathbb{R}^{(L + T) \times D},
\end{equation}
where $W_b$ and $b_b$ are the parameters of a fully connected layer.

(c) Multimodal Cross‐Attention (AlignDiT):
Lastly, instead of fusing the audio-video and text features at the input level, we design the model to gradually incorporate text information while preserving the natural synchronization between audio and video. To this end, we revise each DiT block by inserting a cross-attention layer in addition to self-attention. In this setup, the audio-video representation $h_{av}$ is used as the query, while the text embedding $h_{text}$ serves as the key and value in a multi-head cross-attention mechanism as follows:
\begin{equation}
h=\text{MHCA}(h_{av} W_Q, h_{text} W_K, h_{text} W_V) \in \mathbb{R}^{T \times D},
\end{equation}
where $W_Q$, $W_K$, and $W_V$ are learnable projection matrices. 
By analyzing each approach through (a)-(c) in Section~\ref{subsec:ablation}, we observe that (c) naturally aligns audio-visual features while effectively incorporating text features, achieving the best multimodal alignment. Therefore, we adopt this variant in our AlignDiT model.

\subsection{Training Objective} 
Based on the fused multimodal representations, we train our AlignDiT with multi-task learning. First, we adopt the conditional flow matching (CFM) training objective, which has been proven its effectiveness in generating high-quality data samples in an efficient manner~\cite{lipman2023flow}. The CFM seeks to match a probability path $p_t$ from a tractable distribution $p_0$ to $p_1$ approximating the target distribution. Given the fused multimodal representation $h$ and a noisy mel-spectrogram $x_t$ ($t \in [0, 1]$), AlignDiT is trained to regress the vector field $u_t$ with CFM objectives:
\begin{equation}
    \mathcal{L}_{CFM} = \mathbb{E}_{t,p_t} \|v_t(x_t|h, \theta) - u_t(x_t)\|^2,
\end{equation}
where $\theta$ defines the DiT blocks and $v_t(x_t|h, \theta)$ denotes the estimated vector fields with 
$x_t \sim p_t(x)$.

Second, while the CFM loss encourages natural mel-spectrogram generation, relying solely on the CFM loss can be insufficient for modality alignment, as it provides only an indirect learning signal~\cite{choi2025accelerating}. To address this, we introduce a Connectionist Temporal Classification (CTC)~\cite{graves2006connectionist} loss to guide the intermediate representations of DiT blocks to align more directly with the textual content. Specifically, we attach lightweight projection heads to several intermediate blocks to predict the text sequence from their hidden representations. The loss can be expressed as:
\begin{equation}
\mathcal{L}_\text{CTC} = -\sum_{i \in \mathcal{I}} \log p_{\text{CTC}}(\text{text} \mid h^i),
\end{equation}
where $h^i$ is the output of the $i$-th DiT block and $\mathcal{I}$ denotes the selected layers. The CTC loss encourages the model to retain more linguistic information, without employing external alignment modules.
The total multi-task loss is defined as follows:
\begin{equation}
\mathcal{L}_\text{total}=\mathcal{L}_\text{CFM} + \lambda_\text{CTC}\mathcal{L}_\text{CTC},
\end{equation}
where $\lambda_\text{CTC}$ is a balancing hyperparameter.

\subsection{Audio-only Pretraining}
Generating high-quality speech and learning multimodal alignment simultaneously can be challenging. To provide a better initialization for AlignDiT, we employ an audio-only pretraining phase before conditioning on video and text. The model learns to predict masked regions of the input mel-spectrogram from unmasked context during pretraining, and this has demonstrated its effectiveness in text‐to‐speech task~\cite{liu2024generative}. In addition, we follow the approach~\cite{yu2025representation} that improves convergence speed by distilling rich features into the intermediate layers of DiT, using a self-supervised speech model~\cite{hsu2021hubert} as the teacher. By allowing the model to acquire unconditional speech generation abilities prior to multimodal conditioning, this pretraining phase simplifies the subsequent audio-video-text fusion. Additionally, since audio‐only data is abundant compared to audio-video-text paired data, our audio‐only pretraining strategy is readily scalable and offers a practical way to improve the performance.

\subsection{Multimodal Classifier-Free Guidance}
In diffusion-based generative models, classifier-free guidance~\cite{ho2021classifier} is well-explored to strengthen the influence of conditioning signals during inference. This is achieved by using both conditional and unconditional predictions from the same model to guide the generation process as follows:
\begin{equation}
v_{t,CFG} = v_t(x_t, h) + s \cdot (v_t(x_t, h) - v_t(x_t, \emptyset)),
\end{equation}
where $s$ is guidance scale.

Since each modality exhibits different characteristics, we hypothesize that using a single guidance scale for all modalities may be sub-optimal. To allow better control over each modality during inference, we propose multimodal classifier-free guidance by assigning modality-specific guidance scales:
\begin{equation}
\begin{split}
v_{t,CFG} = &\: v_t(x_t, h_{text}, h_{video}) \\ &+ s_{video} \cdot (v_t(x_t, h_{text}, h_{video}) - v_t(x_t, h_{text}, \emptyset)) \\ &+ s_{text} \cdot (v_t(x_t, h_{text}, \emptyset) - v_t(x_t, \emptyset, \emptyset)),
\end{split}
\end{equation}
where $s_{text}$ is guidance scale for text modality and $s_{video}$ for video. By adjusting $s_{text}$ and $s_{video}$, we can adaptively control the focus between modalities. Higher $s_{text}$ encourages the model to follow the text more closely, improving intelligibility, while higher $s_{video}$ leads to better lip synchronizations.

To support CFG, we apply modality dropout during training by randomly dropping text, video, or both. This not only enables multimodal CFG but also improves robustness in cases where a modality may be missing.

\section{Experiments}
\subsection{Datasets}
We train and evaluate our proposed AlignDiT on the large-scale audio-visual dataset LRS3~\cite{afouras2018lrs3}, which contains 439 hours of English sentence-level data sourced from TED and TEDx talk videos from thousands of speakers. Each video segment contains unconstrained audio-visual speech paired with an accurate transcript, making the dataset suitable for our multimodal setting. Around 131,000 utterances are utilized for training. For evaluation, we construct an LRS3-cross test set consisting of triplets of \{reference speech, text, silent video\}, where the reference speech is from a different utterance by the same speaker, rather than using the ground-truth speech as the reference. This prevents the model from accessing ground-truth speech during testing, ensuring a rigorous evaluation without information leakage. For video-to-speech, instead of using the provided text from the dataset, we utilize an off-the-shelf lip reading model to extract transcripts directly from the silent videos.

\subsection{Evaluation Metrics}
\noindent\textbf{Subjective metrics.}
Since our primary focus is generating natural and synchronized speech, subjective evaluation is essential for accurately assessing model performance~\cite{maiti2023speechlmscore,ahn2024voxsim}. Thus, we conduct human evaluations using Mean Opinion Scores (MOS) for the ADR task along two criteria: naturalness, evaluating the overall quality of the speech; and similarity, assessing speaker similarity between the reference speech and synthesized speech. 20 participants rate the randomly sampled 30 utterances using a 5-point Likert scale, where 1 indicates ``very poor'' (or ``very different'') and 5 indicates ``very good'' (or ``very similar'').

\noindent\textbf{Objective metrics.}
We employ comprehensive objective metrics to evaluate the synthesized speech. For both ADR and video-to-speech tasks, we report Word Error Rate (WER) to assess content accuracy, using Whisper-large-v3~\cite{radford2023robust} to transcribe the synthesized speech. Speaker similarity (spkSIM) is measured by computing the cosine similarity between speaker embeddings extracted from synthesized and reference speech using a WavLM-large-based speaker verification model~\cite{chen2022wavlm}. 
Lip-sync accuracy (AVSync) between the video and synthesized speech is measured using AVHubert~\cite{shi2022learning:avhubert}. Specifically, we compute the cosine similarity between AVHubert features extracted from video paired with ground-truth speech and features extracted from video paired with synthesized speech. This assessment has been shown in \cite{yaman2024audio} to be more robust and effective for validating audio-visual synchronization compared to conventional lip-sync accuracy metrics based on SyncNet~\cite{syncnet}.
For the alignment task, we evaluate alignment accuracy by comparing word-level timestamps obtained from our method with ground truth timestamps derived from original audio alignment. We report the average absolute time error per word (in milliseconds).

\begin{table}[!t]
    \centering
    \caption{Ablation study of modality conditioning method.}
    \vspace{-3mm}
    \label{tab:arch}
    \resizebox{0.79\linewidth}{!}{\begin{tabular}{cc|ccc}
    \toprule
    \textbf{Method} & $\mathcal{L}_{CTC}$ & WER $\downarrow$ & AVSync $\uparrow$ & spkSIM $\uparrow$ \\ 
    \midrule
    \multirow{2}{*}{(a)} & \xmark & 28.828 & 0.666 & 0.530 \\
    & \cmark & 2.236 & 0.748 & 0.511 \\
    \midrule
    \multirow{2}{*}{(b)} & \xmark & 5.456 & 0.726 & 0.536 \\
    & \cmark & 1.917 & 0.745 & 0.519 \\
    \midrule
    \multirow{2}{*}{(c)} & \xmark & 2.507 & 0.745 & \textbf{0.543} \\
    & \cmark & \textbf{1.401} & \textbf{0.751} & 0.515 \\
    \bottomrule
    \end{tabular}}
    \vspace{-2mm}
\end{table}


\begin{table}[!t]
    \centering
    \caption{Ablation study of audio-only pretraining.}
    \vspace{-3mm}
    \label{tab:pretrain}
    \resizebox{0.81\linewidth}{!}{\begin{tabular}{l|ccc}
    \toprule
    \textbf{Pretraining data.} & WER $\downarrow$ & AVSync $\uparrow$ & spkSIM $\uparrow$ \\ 
    \midrule
    \xmark & 2.040 & 0.746 & 0.396 \\
    LRS3 & 1.720 & 0.750 & 0.503 \\
    LibriSpeech & \textbf{1.401} & \textbf{0.751} & \textbf{0.515} \\
    \bottomrule
    \end{tabular}}
    \vspace{-3mm}
\end{table}


\subsection{Implementation Details}
\noindent\textbf{Data preprocessing.} For audio features, we use 16~kHz mono audio and convert it into 80 bins mel-spectrogram using a filter size of 640, a hop size of 160, resulting in a frame rate of 100~Hz. We utilize 25~fps video and extract visual features as follows. Face detection is performed using RetinaFace~\cite{deng2020retinaface}, followed by facial landmark extraction using FAN~\cite{bulat2017far}. We crop the lip-centered region based on the detected landmarks and resize it into 88~×~88. The pretrained AV-HuBERT (Large) model~\cite{shi2022learning:avhubert} is utilized to extract 25 Hz visual representations, which result in a fixed 1:4 length ratio to the audio features. For text, we represent input as a sequence of characters. Compared to the audio features, the character sequence generally has a shorter temporal length~\cite{ren2019fastspeech}.

\noindent\textbf{Architecture.}
The visual feature encoder is composed of two transposed convolution layers with stride 2, followed by two Conformer~\cite{gulati2020conformer} encoder layers, each with embedding size of 512, 4 attention heads, and a 1024-dimensional feed-forward layer. For the text encoder, we adopt 4 layers of ConvNeXt v2~\cite{woo2023convnext} with 512-dimensional hidden embeddings. After concatenating multimodal features, we apply a linear layer to project the fused representation. This representation is then processed by 18 DiT blocks, each with 768-dimensional embedding size, 12 attention heads, and a 3072-dimensional feed-forward layer.

\noindent\textbf{Training.} We train AlignDiT using AdamW~\cite{loshchilov2019decoupled} optimizer with a warmup of 20k steps to a peak learning rate of $7.5 \times 10^{-5}$, followed by linearly decay. Pretraining is conducted for 500k steps on audio-only data with a total batch size of 0.3 hours. We finetune the model for 400k steps using paired audio, video, and text data with a total batch size of 0.1 hours. $\lambda_{CTC}$ is set to 0.1 to balance the CTC loss against the CFM loss in initial stage. The modality dropout probability is set to 0.2 for text, video, and all modalities, respectively.

\noindent\textbf{Inference.}
During inference, we apply Exponential Moving Averaged (EMA) weights with a decay rate of 0.999 to stabilize the model prediction. AlignDiT takes a reference speech along with its corresponding transcript as inputs, and the total duration is determined by the input video length. For sampling, we use the Euler ODE solver with timestep scaling based on sway sampling strategy with a coefficient of -1, following F5-TTS~\cite{chen2025f5}. To convert generated mel spectrograms into waveforms, we employ a HiFi-GAN~\cite{kong2020hifi} model trained on the LRS3 dataset.

\subsection{Baseline Models}
We compare AlignDiT with state-of-the-art open-source ADR systems, HPMDubbing~\cite{cong2023learning} and StyleDubber~\cite{cong2024styledubber}. To ensure a fair comparison and improve their generalization capability, we train these models on the LRS3 dataset~\cite{afouras2018lrs3}, which is significantly larger than the datasets originally used in their training. Additionally, we replace the lip feature extractor in each model with AV-HuBERT (large)~\cite{shi2022learning:avhubert}, unifying the lip feature extractor across AlignDiT, HPMDubbing, and StyleDubber to purely evaluate and compare their effectiveness under identical settings. Note that both baseline models require an explicit duration aligner during training.
For video-to-speech, we compare our method with DiffV2S~\cite{choi2023diffv2s}, Intelligible~\cite{choi2023intelligible}, and LipVoicer~\cite{yemini2024lipvoicer}, which are specifically designed for this task.
For visual forced alignment, we compare against visual keyword spotting methods such as KWS-Net~\cite{momeni2020seeing} and Transpotter~\cite{prajwal2021visual}, as well as CTC-based methods~\cite{kurzinger2020ctc} and DVFA~\cite{kim2023deep}.

\begin{table}[!t]
    \centering
    \caption{Ablation study of multimodal CFG.}
    \vspace{-3mm}
    \label{tab:cfg}
    \resizebox{0.72\linewidth}{!}{\begin{tabular}{cc|ccc}
    \toprule
    $s_{text}$ & $s_{video}$ & WER $\downarrow$ & AVSync $\uparrow$ & spkSIM $\uparrow$ \\ 
    \midrule
    0 & 0 & 3.785 & 0.716 & 0.391 \\
    2 & 2 & 2.310 & 0.758 & 0.497 \\
    5 & 2 & \textbf{1.401} & 0.751 & \textbf{0.515} \\
    5 & 5 & 2.507 & \textbf{0.760} & 0.501 \\
    \bottomrule
    \end{tabular}}
    \vspace{-2mm}
\end{table}


\begin{table}[!t]
    \centering
    \caption{Ablation study of input modalities.}
    \vspace{-3mm}
    \label{tab:modality}
    \resizebox{0.72\linewidth}{!}{\begin{tabular}{cc|ccc}
    \toprule
    \multicolumn{2}{c|}{\textbf{Input mod.}} & \multirow{2}{*}{WER $\downarrow$} & \multirow{2}{*}{AVSync $\uparrow$} & \multirow{2}{*}{spkSIM $\uparrow$} \\
    text & video & \\
    \midrule
    \xmark & \cmark & 27.820 & 0.674 & 0.486 \\
    \cmark & \xmark & 1.769 & 0.270 & 0.501 \\
    \cmark & \cmark & \textbf{1.401} & \textbf{0.751} & \textbf{0.515} \\
    \bottomrule
    \end{tabular}}
    \vspace{-3mm}
\end{table}


\begin{figure*}[ht]
    \centering
    \includegraphics[width=1.0\linewidth]{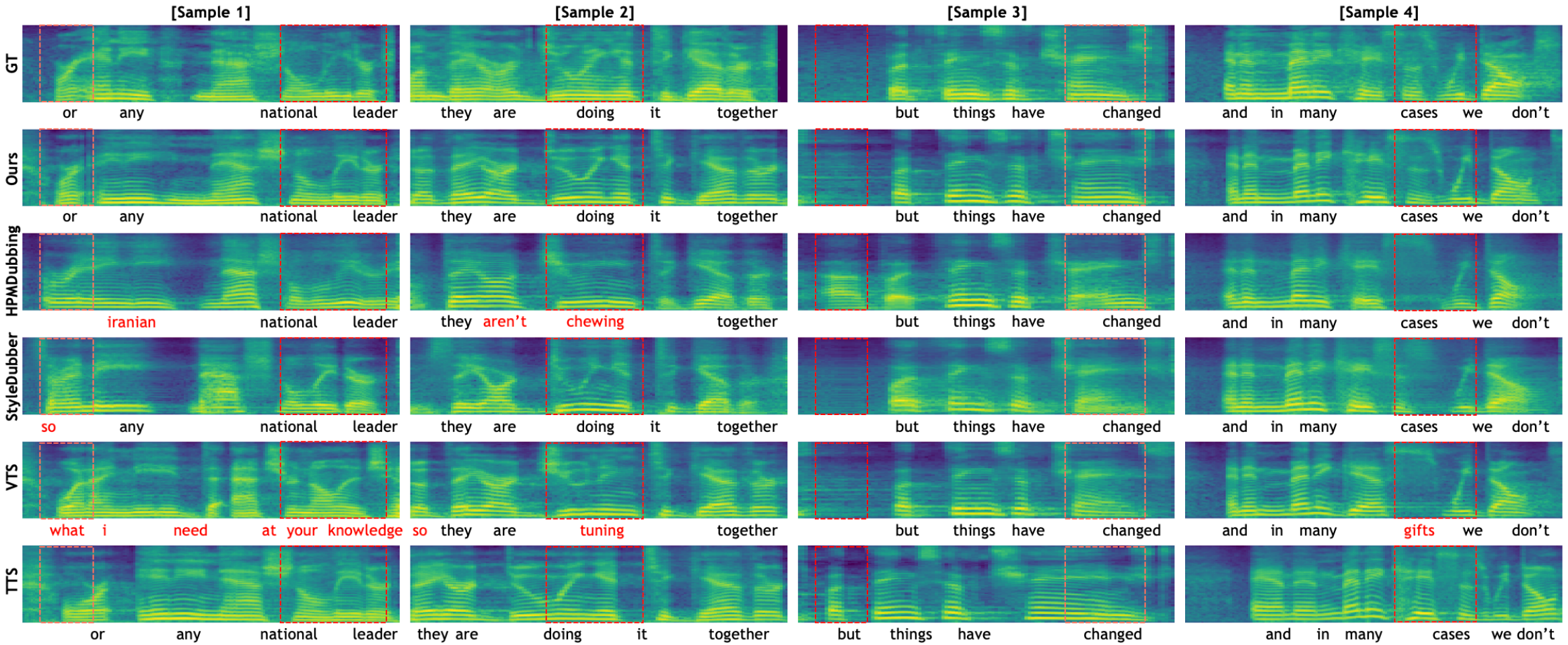}
    \vspace{-5mm}
    \caption{We qualitatively compare mel-spectrogram visualizations of ground-truth speech and synthesized speech from the ADR task using AlignDiT (ours) and existing methods, HPMDubbing and StyleDubber. We also provide results from unimodal inputs, either video-only (VTS) or text-only (TTS). The text below each mel-spectrogram represents time-aligned speech extracted using Whisper-large-v3, with red text indicating incorrectly synthesized word compared to the ground truth.} 
    \label{fig:qualitative}
    \vspace{-1mm}
\end{figure*}

\begin{table*}[!t]
    \centering
    \caption{Experimental results on LRS3-cross set. Subjective MOS results are presented with 95\% confidence interval. 
    $\uparrow$ represents that a higher score is better, and $\downarrow$ denotes that a lower score is better. 
    The best-performing result is shown in bold.
    }
    \vspace{-2mm}
    \label{tab:main}
    \resizebox{0.66\linewidth}{!}{\begin{tabular}{l|cc|ccc}
    \toprule
    \multirow{2}{*}{\textbf{Method}} & \multicolumn{2}{c|}{Subjective} & \multicolumn{3}{c}{Objective} \\ 
    & Naturalness $\uparrow$ & Speaker Similarity $\uparrow$ & WER $\downarrow$ & AVSync $\uparrow$ & spkSIM $\uparrow$  \\ 
    \midrule
    GT &4.13$\pm$0.13 &3.96$\pm$0.14 & 2.335 & 1.000 & 0.562  \\ \cmidrule{1-6}
    HPMDubbing~\cite{cong2023learning} &2.37$\pm$0.14 &2.43$\pm$0.16 & 5.382 & 0.750 & 0.287  \\
    StyleDubber~\cite{cong2024styledubber} &1.79$\pm$0.12 &2.35$\pm$0.18 & 3.170 & 0.586 & 0.349  \\
    \textbf{Ours} &{\bf 3.79$\pm$0.16} &{\bf 3.96$\pm$0.13} & \textbf{1.401} & \textbf{0.751} & \textbf{0.515} \\
    \bottomrule
    \end{tabular}}
    \vspace{-2mm}
\end{table*}


\section{Results}
\subsection{Ablation Studies}
\label{subsec:ablation}
We analyze the contribution of each AlignDiT component through in-depth ablation studies. We conduct a series of experiments using various objective metrics, \emph{i.e.} WER, AVSync, and spkSIM, and present comprehensive findings on modality conditioning method, audio-only pretraining, multimodal CFG, and input modalities.

\noindent\textbf{Modality conditioning method.}
To validate the effectiveness of the conditioning strategy of AlignDiT, we carry out a detailed analysis of alternative modality conditioning strategies.
We compare three conditioning variants---(a) early fusion \& self-attention, (b) prefix \& self-attention, and (c) multimodal cross-attention---and analyze the impact of the CTC loss ($\mathcal{L}_{\mathrm{CTC}}$) in each case.
The results in Table~\ref{tab:arch} confirm the strength of the proposed conditioning method (multimodal cross-attention) which achieves the best performance in WER, AVSync, and spkSIM.

In addition, the absence of $\mathcal{L}_{CTC}$ leads to consistent quality degradation across all variants and metrics, except for a slight deviance in spkSIM of the (c) multimodal cross-attention.
In particular, the WER is significantly worsened in every variant, indicating that applying $\mathcal{L}_{CTC}$ facilitates more accurate alignment learning between the input text and output speech within the DiT blocks.

\noindent\textbf{Audio-only pretraining.}
Table~\ref{tab:pretrain} illustrates the benefits of audio-only pretraining. Pretraining AlignDiT on the LRS3 dataset, which is also used during the main training phase, shows consistent quality improvements across all evaluation metrics. Even greater improvements are achieved when using LibriSpeech~\cite{panayotov2015librispeech}, a dataset not seen during the subsequent training. These results underscore the benefits of leveraging diverse audio data during pretraining, and demonstrate that the performance of AlignDiT can be readily enhanced using easily accessible audio resources.

\noindent\textbf{Multimodal CFG.}
In order to explore how AlignDiT benefits from its multimodal CFG, we experiment with different values of $s_{text}$ and $s_{video}$. From the results in Table~\ref{tab:cfg}, we derive two key findings. First, applying CFG ($s_{text} = s_{video} \in \{2, 5\}$) consistently improves performance compared to not using CFG ($s_{text} = s_{video} = 0$), aligning with observations from previous works~\cite{ho2021classifier, le2023voicebox}. More importantly, leveraging the proposed multimodal CFG, i.e., adjusting $s_{text}$ and $s_{video}$ to balance information from each modality, further enhances overall performance, demonstrating its effectiveness in synthesizing more natural speech. Through analysis of optimal guidance scales for each modality, we set $s_{text}$ = 5 and $s_{text}$ = 2, which achieves the best quality in terms of WER and spkSIM.

\noindent\textbf{Input modalities.}
To examine the effect of multimodal inputs, we compare models trained with video only, text only, and both text and video modalities (Table~\ref{tab:modality}). The WER significantly increases when text input is omitted, highlighting the essential role of textual information in generating accurate and intelligible speech. Excluding the video modality also degrades overall performance,  particularly in terms of AVSync. In contrast, utilizing both modalities yields speech that is not only intelligible but also well-synchronized, underscoring the advantages of multimodal-to-speech generation.

\subsection{Quantitative Comparison}
Table~\ref{tab:main} shows both subjective and objective evaluation results of AlignDiT compared to baseline systems. As shown, AlignDiT consistently outperforms all baselines across all evaluation metrics. To be specific, in the subjective evaluation, our method achieves a naturalness score of 3.79 and a speaker similarity score of 3.96, substantially surpassing the baselines by a large margin. These results suggest that AlignDiT generates fluent speech that is perceptually superior in both naturalness and speaker similarity.

The objective evaluation further supports the effectiveness of AlignDiT. It achieves the lowest WER of 1.401, demonstrating its ability to accurately synthesize the input text. It also obtains the highest AVSync score of 0.751. This indicates AlignDiT effectively leans audio-video temporal alignment, without depending on additional aligners or duration predictors which commonly used in the baselines. In terms of speaker consistency, AlignDiT also achieves the best spkSIM score, demonstrating its ability to effectively mimic the voice characteristics of the target speaker.

\begin{table}[!t]
    \centering
    \caption{Video-to-Speech benchmark.}
    \vspace{-2mm}
    \label{tab:v2s}
    \resizebox{0.80\linewidth}{!}{\begin{tabular}{l|ccc}
    \toprule
    \textbf{Method} & WER $\downarrow$ & AVSync $\uparrow$ & spkSIM $\uparrow$ \\ 
    \midrule
    DiffV2S~\cite{choi2023diffv2s} & 35.210 & 0.608 & 0.115 \\
    Intelligible~\cite{choi2023intelligible} & 27.432 & 0.675 & 0.316 \\
    LipVoicer~\cite{yemini2024lipvoicer} & 21.164 & 0.524 & 0.094 \\
    Ours & \textbf{19.513} & \textbf{0.688} & \textbf{0.508} \\
    \bottomrule
    \end{tabular}}
\end{table}


\subsection{Qualitative Comparison}
{We visually compare the mel-spectrograms converted from synthesized speech produced by existing ADR methods~\cite{cong2023learning,cong2024styledubber} and our AlignDiT with multimodal input, alongside those from ground-truth speech, in Fig.~\ref{fig:qualitative}. We also include results synthesized from unimodal inputs: video-only (VTS) and text-only (TTS). Focusing on the orange and red boxes in columns 1 and 2, as well as the transcribed speech below each mel-spectrogram, we observe that existing methods frequently generate incorrect speech and blurry spectrograms. In columns 3 and 4, although existing methods produce accurate speech content, they occasionally generate unintended sounds or produce mel-spectrograms that lack details and appear blurry. Regarding VTS and TTS, we observe consistent results in Sec.~\ref{subsec:ablation}. For VTS across all samples, since no text input is provided, content accuracy becomes significantly unstable, leading to incorrect speech. Conversely, TTS generates accurate speech content across all examples due to the text input, yet the absence of video input results in temporal misalignment compared to the ground-truth. In contrast, our model consistently synthesizes accurate speech content, exhibiting fine details in the mel-spectrogram that closely match the ground-truth across all examples. These results clearly demonstrate that our model effectively synthesizes high-quality, accurate speech comparable to the ground-truth.}

\subsection{Applications}
Our proposed AlignDiT is robust and flexible across diverse multimodal scenarios. We demonstrate the versatility of our method by showing its effective generalization to related multimodal tasks, such as video-to-speech synthesis and visual forced alignment.

\noindent\textbf{Video-to-speech.}
Unlike ADR, video-to-speech takes silent video (without text) as input to read lip movements and synthesize corresponding speech. Some prior works~\cite{choi2023diffv2s,choi2023intelligible} operate in textless manner, while others, such as LipVoicer~\cite{yemini2024lipvoicer}, leverage off-the-shelf lip reading models to guide speech synthesis. For a fair comparison with LipVoicer, we adopt the same lip reading model~\cite{ma2023auto} to obtain pseudo-text labels. Table~\ref{tab:v2s} summarizes the performance comparison between our approach and existing methods. Interestingly, our model significantly outperforms existing methods across all evaluation metrics, including LipVoicer, which is specifically designed for this task and also utilizes an expert lip reading model. This highlights the robustness of our multimodal alignment approach, demonstrating that it generalizes effectively and achieves superior performance even on tasks beyond its primary design.

\begin{table}[!t]
    \centering
    \caption{Visual forced alignment benchmark.}
    \vspace{-2mm}
    \label{tab:vfa}
    \resizebox{0.58\linewidth}{!}{\begin{tabular}{l|cc}
    \toprule
    \textbf{Method} & MAE $\downarrow$ & ACC $\uparrow$ \\ 
    \midrule
    KWS-Net~\cite{momeni2020seeing} & 262.9ms & 42.6\% \\
    CTC-based~\cite{kurzinger2020ctc} & 124.5ms & 60.6\% \\
    Transpotter~\cite{prajwal2021visual} & 167.3ms & 61.8\% \\
    DVFA~\cite{kim2023deep} & 97.7ms & 80.2\% \\
    Ours & \textbf{41.5ms} & \textbf{83.7\%} \\
    \bottomrule
    \end{tabular}}
\end{table}


\noindent\textbf{Visual forced alignment (VFA).}
Conventionally, the VFA task involves identifying timestamps for each word or phoneme in silent videos. Since our model can synthesize speech given silent video and text input, we bypass direct comparison of silent video and text, and instead leverage these inputs to generate speech signals. We then apply the Montreal Forced Aligner~\cite{mcauliffe2017montreal} to align the synthesized speech with the corresponding text, thus determining timestamps of each word for VFA task. It is worth noting that we use a single canonical reference speech across all test samples, removing the need for speaker-specific references. Table~\ref{tab:vfa} presents a comparison of alignment performance on the LRS3 dataset. The synthesized speech from our proposed AlignDiT is highly synchronized with the input video, enabling precise forced alignment with text, resulting in significantly better performance compared to existing methods~\cite{momeni2020seeing, prajwal2021visual,kurzinger2020ctc,kim2023deep} specifically designed for this task. These results support that our model generates highly accurate and temporally synchronized speech from multimodal inputs.

\section{Conclusion}
We introduced AlignDiT, a unified framework for generating accurate, natural, and synchronized speech from text, video, and reference audio. Through extensive analysis, we explore various configurations and identify the most effective strategy for aligning multiple modalities, without the need for explicit duration modeling. We also proposed a multimodal classifier-free guidance mechanism that adaptively balances information across modalities. AlignDiT achieves state-of-the-art performance across several benchmarks and demonstrates its effectiveness in key multimodal tasks, including video-to-speech synthesis and visual forced alignment. We believe our findings offer valuable insights for future research in multimodal alignment and generation.

\begin{acks}
This work was supported by IITP grants funded by the Korean government (MSIT, RS-2025-02263169, Detection and Prediction of Emerging and Undiscovered Voice Phishing and RS-2024-00457882, National AI Research Lab Project).
\end{acks}

\bibliographystyle{ACM-Reference-Format}
\bibliography{main}

\end{document}